\documentclass{article}

\usepackage[final]{neurips_data_2024}
\usepackage[T1]{fontenc}
\usepackage{microtype}
\usepackage{graphicx}
\usepackage{booktabs}
\usepackage{siunitx}
\usepackage{subcaption}
\usepackage{hyperref}

\usepackage{booktabs}
\usepackage{siunitx}
\DeclareSIUnit{\fps}{fps}

\usepackage[utf8]{inputenc} 
\usepackage[T1]{fontenc}    
\usepackage{hyperref}       
\usepackage{url}            
\usepackage{booktabs}       
\usepackage{amsfonts}       
\usepackage{nicefrac}       
\usepackage{microtype}      
\usepackage{xcolor}         
\usepackage{booktabs}       
\usepackage{tikz}           
\usepackage{siunitx}  
\usepackage{amsmath}  
\usepackage{array}    
\usepackage{microtype} 

\usepackage{microtype} 
\usepackage{subcaption}
\usepackage{listings}
\usepackage{xcolor}
\usepackage{graphicx}
\title{Look and Tell: A Dataset for Multimodal Grounding Across Egocentric and Exocentric Views}

\author{%
  Anna Deichler \\
  TMH \\
  KTH Royal Institute of Technology \\
  \texttt{deichler@kth.se} \\
  \And
 Jonas Beskow \\
 TMH \\
  KTH Royal Institute of Technology \\
  \texttt{beskow@kth.se}\\
}

\begin{document}

\maketitle

\begin{abstract}
We introduce Look and Tell, a multimodal dataset for studying referential communication across egocentric and exocentric perspectives. Using Meta Project Aria smart glasses and stationary cameras, we recorded synchronized gaze, speech, and video as 25 participants instructed a partner to identify ingredients in a kitchen. Combined with 3D scene reconstructions, this setup provides a benchmark for evaluating how different spatial representations (2D vs. 3D; ego vs. exo) affect multimodal grounding. The dataset contains 3.67 hours of recordings, including 2,707 richly annotated referential expressions, and is designed to advance the development of embodied agents that can understand and engage in situated dialogue.
\end{abstract}

\section{Introduction}
Understanding how humans coordinate gaze, gestures, and speech in referential communication 
is critical for building embodied agents that can interact naturally and exhibit spatial intelligence. 
While gaze has long been recognized as a cue for attentional focus, most research has emphasized its 
temporal relationship to speech in controlled tasks. Less attention has been given to how spatial representations of the 
environment---whether in 2D image space, 3D reconstructions, or across egocentric and exocentric 
perspectives---influence multimodal spatial grounding.

We present a new dataset collected with 25 participants in a naturalistic kitchen setting, where 
participants recalled recipe ingredients while wearing Meta Aria smart glasses \citep{Engel2023ProjectAria}. The glasses provided 
synchronized gaze and speech streams, complemented by exocentric GoPro recordings. This dual-view 
setup enables both fine-grained analysis of gaze--speech synchrony and systematic evaluation of 
representation choices in multimodal grounding. Specifically, the dataset supports comparisons between 
2D and 3D scene representations, as well as between egocentric and exocentric perspectives, offering a 
unique testbed for investigating how humans and embodied models organize spatial knowledge during 
communication. This is crucial for human--robot interaction, where an agent must understand references 
from a user’s point of view and integrate multimodal signals such as gaze, gestures, and speech. This dual-view paradigm is critical for tackling challenges in \textbf{shared autonomy and human-robot collaboration}, where an agent must simultaneously interpret a user's first-person intent while maintaining an objective, third-person model of the world state.

Our contributions are threefold: (1) a multimodal dataset of synchronized gaze, speech, and dual-view video designed to study \textbf{situated dialogue}; (2) an annotation pipeline for \textbf{aligning spatial concepts across language and vision}; and (3) a new \textbf{benchmark for evaluating spatial intelligence} in \textbf{grounded communication}. By capturing rich, natural behavior, our dataset provides a foundation for future analysis of how gesture, gaze, and speech jointly enable \textbf{multimodal spatial grounding}.
\section{Related Work}

\textbf{Gaze and speech in communication.} Prior work shows that gaze strongly predicts communicative goals \citep{hanna2007speakers,brennan2007coordinating}, and has been combined with gestures or speech to improve reference resolution \citep{renner2014gaze}. Event synchronization studies \citep{kaur2003synchronization} highlight the potential of aligning multimodal streams, though mainly in controlled tasks. Recent work emphasizes gaze in instruction and teaching contexts \citep{wagner2023multimodal} and in virtual environments \citep{tanriverdi2000interacting}. However, temporal alignment of gaze and speech in natural referential tasks remains underexplored.

\textbf{Multimodal communication datasets.} Several datasets integrate multiple communicative channels, such as speech, gaze, and gesture in collaborative tasks \citep{kontogiorgos2018multimodal}. VENUS \citep{kim2025venus} combines speech, facial expressions, and body pose. In vision--language research, gaze has been used for captioning \citep{sugano2016seeing}, VQA \citep{vasudevan2018object,ilaslan2023gazevqa}. Despite this progress, datasets combining first-person gaze with concurrent speech in referential tasks are rare.

\textbf{Spatial representations for grounding.} Recent research highlights the importance of spatial representations for multimodal agents. Ego4D and related egocentric datasets emphasize large-scale video and gaze \citep{grauman2022ego4d}, while 3D QA datasets such as ScanQA \citep{gao2021scanqa} and EmbodiedQA \citep{das2018embodiedqa} explore grounding in reconstructed scenes. Yet, few datasets allow direct comparison between 2D vs.\ 3D representations or ego vs.\ exo perspectives in the same task. Our dataset bridges this gap, enabling systematic evaluation of how representational choices affect grounding in situated referential communication.

\section{Data Collection \& Dataset Description}

\subsection{Experiment Design}
\textbf{Participants.} 25 participants (18 female, 7 male; age range 22--37) were recruited at KTH Kitchen Lab.  
\textbf{Materials.} Aria smart glasses (for eye tracking and audio recording), GoPro cameras (environment capture), recipes, and food ingredients.  
\textbf{Task.} Each participant memorized a step of a recipe and then instructed it to a conversational partner. While speaking, they were asked to identify and refer to the corresponding ingredients in the environment.  

\paragraph{Procedure.}
\begin{enumerate}
    \item Participant reads recipe step.
    \item Participant recalls step aloud while locating ingredients.
    \item Egocentric Aria glasses record synchronized gaze, audio, video, and pointcloud data; exocentric GoPro records audio and video feed.
    \item Sessions were recorded across five recipes per participant.
\end{enumerate}

\subsection{Dataset Description}
The \textbf{Look and Tell} dataset, curated by KTH Royal Institute of Technology, offers a unique resource for investigating the interplay between visual attention and spoken language. This dataset was  collected using Aria smart glasses, which allowed for the simultaneous and real-time capture of participants' eye-tracking data and speech audio.

\begin{figure}[t]
    \centering
    \includegraphics[width=0.8\linewidth]{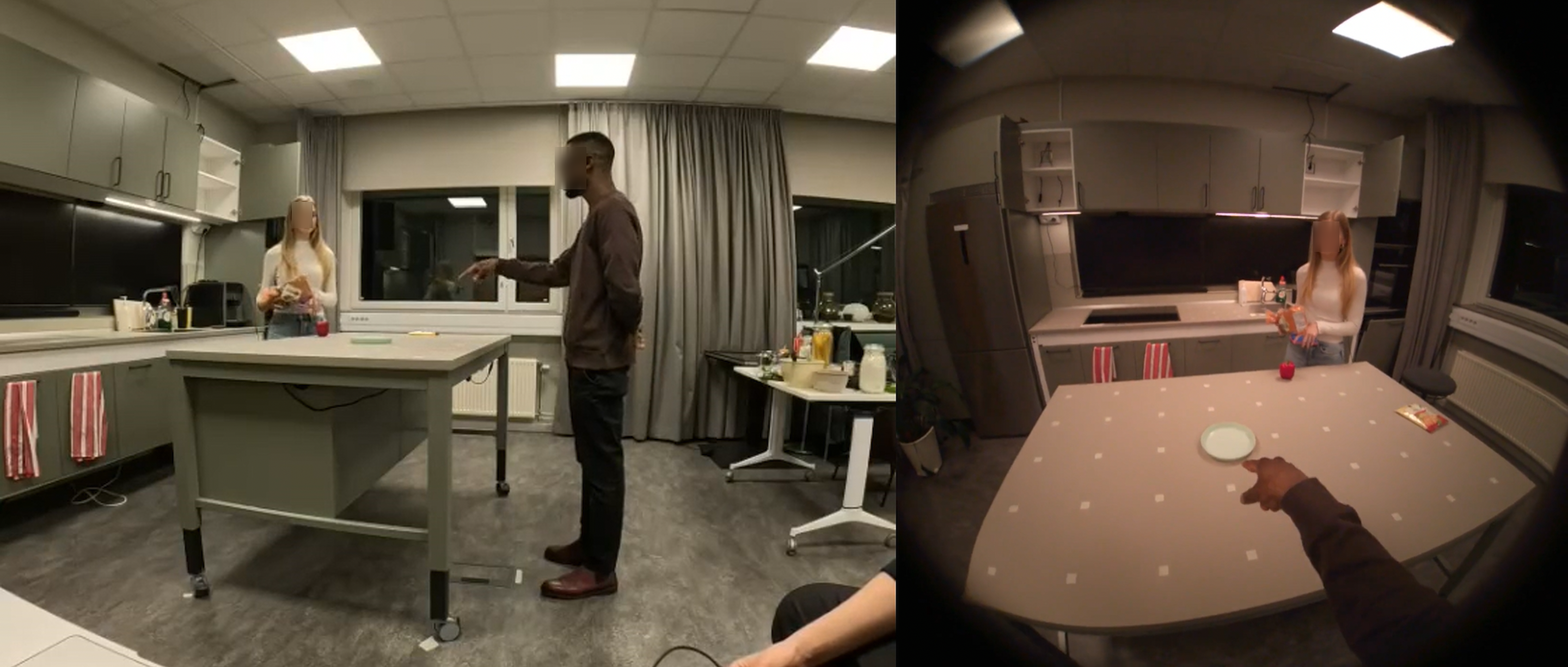}
    \caption{Example of synchronized video feeds. The exocentric (left) view provides situational context, while the egocentric (right) view captures the participant's first-person perspective, including their gaze target (green circle, overlaid for visualization).}
    \label{fig:sync_example}
\end{figure}

The experimental paradigm involved participants identifying various food items from a recipe. This task was designed to elicit natural referential communication, providing rich, ecologically valid data on how individuals visually fixate on objects while verbally describing or identifying them. The recordings from the Aria glasses provide a synchronized stream of high-resolution eye-tracking data and corresponding speech, enabling detailed analysis of gaze-speech synchronization patterns. This is  complemented by exocentric GoPro recordings. This dual-view setup enables fine-grained analysis of gaze–speech synchrony and evaluation of representational choices in multimodal grounding (see Fig.~\ref{fig:sync_example}).

\subsection{Data Modalities and Format}
\begin{itemize}
    \item \textbf{Eye-tracking data:} fixation events, gaze vectors.
    \item \textbf{Speech audio:} raw speech signal, later transcribed and segmented with WhisperX.
    \item \textbf{Video recordings:}
    \begin{itemize}
        \item Egocentric view: raw video from the head-mounted Aria cameras (MP4 format), recorded at up to 1408$\times$1408 pixels and 30\,fps, synchronized with audio and gaze.  
        \item Exocentric view: side-view video recordings from stationary GoPro cameras, providing third-person context of the participant and environment.  
    \end{itemize}
    \item \textbf{Pointcloud representation} of the scene.
    \item \textbf{Recipe and ingredient metadata}, including surface positions.
\end{itemize}

\begin{figure}[t]
    \centering
    \includegraphics[width=0.8\linewidth]{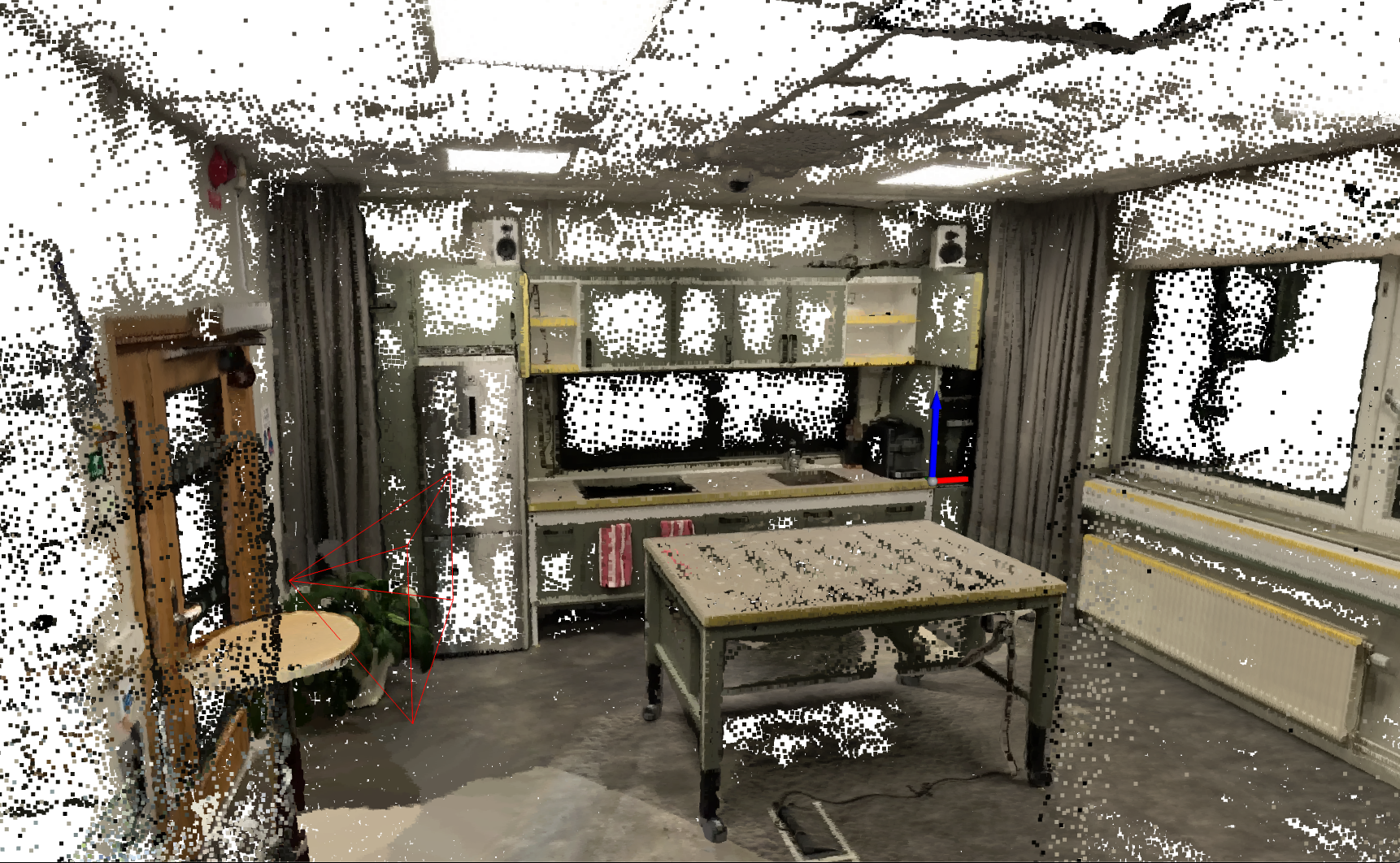}
    \caption{Example of reconstructed 3D room point cloud used as the canonical reference space. All point clouds extracted from Aria recordings are aligned and canonicalized to this shared coordinate system.}
    \label{fig:scene}
\end{figure}

\paragraph{3D Room Reconstruction.} 
In addition to synchronized gaze and speech recordings, we also obtained reconstructed 3D models of the kitchen environment from separate video sessions. Using the Meta Project Aria MPS service \citep{ProjectAriaTools}, we extracted image frames from room recordings and generated point clouds that were then canonicalized into a shared coordinate system (Fig.~\ref{fig:scene}). 
This process aligns multiple reconstructions from different recording sessions, yielding a consistent 3D representation of the environment. 
These reconstructions provide additional spatial context beyond 2D video, enabling research on how embodied agents can exploit 3D scene structure for referential grounding and multimodal interaction.

\subsection{Dataset Statistics}
We recruited 25 participants (aged 22--37; 18 female, 7 male).  
Across 125 sessions, the dataset contains a total of \textbf{396,208 RGB frames} ($\sim$3.67 hours at 30\,fps).  
Sessions lasted on average 1.8\,±\,0.7 minutes (range 0.7--4.3), corresponding to 3,170\,±\,1,329 frames per session (range 1,223--7,730).  
The longest session was 4.3 minutes and the shortest 0.7 minutes. See details on \ref{tab:dataset_summary}.  

Participants reported the following native languages: Korean (1), Indian languages (2), Icelandic (1), Chinese Mandarin (10), Spanish (2), Swedish/English bilingual (1), Swedish (4), Turkish (1), Nigerian (1), Indonesian (1), Portuguese (1), and Russian (1).

\begin{table}[t]
\centering
\caption{KTH-ARIA Referential Dataset Summary Statistics}
\label{tab:dataset_summary}
\begin{tabular}{@{}>{\bfseries}p{0.28\linewidth} p{0.67\linewidth}@{}}
\toprule
Participants / Sessions & 25 participants / 125 unique sessions \\
Total RGB Frames       & \num{396208} frames \\
Total Duration         & \qty{3.67}{\hour} (\qty{220.1}{\minute}) at \qty{30}{\fps} \\
\addlinespace
Frames per Session &
  \begin{tabular}[t]{@{}l@{}}
  Mean: \num{3169.7} $\pm$ \num{1328.9} \\
  Range: \num{1223}--\num{7730}
  \end{tabular} \\
\addlinespace
Duration per Session &
  \begin{tabular}[t]{@{}l@{}}
  Mean: \qty{1.8}{\minute} $\pm$ \qty{0.7}{\minute} \\
  Range: \qty{0.7}{\minute}--\qty{4.3}{\minute}
  \end{tabular} \\
\bottomrule
\end{tabular}
\end{table}

\section{Annotation Pipeline and Analysis}
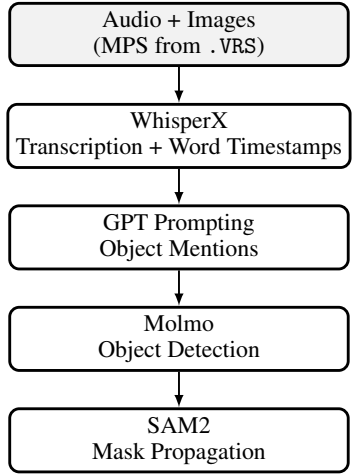
\begin{figure}[t]
\centering
\scalebox{0.9}{ 
\begin{tikzpicture}[
  node distance=1.5cm, 
  >=latex,
  box/.style={
    rectangle,
    rounded corners,
    draw=black,
    very thick,
    minimum height=0.9cm,
    minimum width=5cm,
    align=center
  }
]

\node[box, fill=gray!10] (audio) {Audio + Images \\ (MPS from \texttt{.VRS})};
\node[box, below of=audio] (whisperx) {WhisperX \\ Transcription + Word Timestamps};
\node[box, below of=whisperx] (gpt) {GPT Prompting \\ Object Mentions};
\node[box, below of=gpt] (molmo) {Molmo \\ Object Detection};
\node[box, below of=molmo] (sam2) {SAM2 \\ Mask Propagation};

\draw[->, thick] (audio) -- (whisperx);
\draw[->, thick] (whisperx) -- (gpt);
\draw[->, thick] (gpt) -- (molmo);
\draw[->, thick] (molmo) -- (sam2);

\end{tikzpicture}}
\caption{Annotation pipeline for the egocentric camera: synchronized audio and images are processed with WhisperX for word-level transcripts, GPT for object mentions, Molmo for object detection, and SAM2 for mask propagation.}
\label{fig:annotation_pipeline}
\end{figure}

To enable fine-grained analysis of gaze--speech synchrony, we developed a multi-stage annotation pipeline that integrates audio transcription, language-based reference extraction, and multimodal object detection.  

\paragraph{Audio acquisition.} 
Audio data was obtained from the \textbf{Meta Project Aria MPS service}, which allowed us to request synchronized audio streams from the original \texttt{.vrs} files alongside the corresponding image frames. The images were pre-processed to remove fisheye distortion, yielding undistorted RGB frames aligned with the audio.

\paragraph{Speech transcription.} 
We transcribed the audio using \textbf{WhisperX} \citep{bain2023whisperx}, which provides robust automatic speech recognition together with \textbf{word-level timestamps}. These timestamps form the backbone of our temporal alignment between speech and other modalities.

\paragraph{Reference extraction.} 
Using the transcribed text, we employed \textbf{GPT-based prompting}  to identify mentions of ingredients and distractor objects referenced in the recipe instructions, as well as additional objects spontaneously mentioned by participants \citep{openai2023gpt4}. Here, we use \textit{additional object} to denote non-ingredient items from the recipe/distractor lists 
that participants sometimes referred to explicitly (e.g., \emph{fridge}, \emph{sink}, \emph{stove}).  This step produced a set of candidate \textbf{mentions} linked to precise word spans in the transcripts. The full prompting templates used for this step are provided in the Appendix.  
\subsection{GPT-based Mention Linking}
\label{sec:mention_linking}

To process raw speech transcripts into structured data for multimodal analysis, we automatically convert WhisperX word-level transcripts into span-level ingredient, utensil, and object mentions using a hybrid pipeline (Fig.~\ref{fig:annotation_pipeline}). This process leverages a large language model augmented with deterministic rules to ensure consistency.

\paragraph{Inputs.} The pipeline uses three sources of information: (a) tokenized transcripts with word-level timings from WhisperX, (b) recipe metadata and ingredient lists from \texttt{recipes.json}, and (c) a pre-defined set of scene distractors (which are ingredients not used in current recipe).

\paragraph{LLM Labelling.} We prompt a GPT model with the lower-cased transcripts and a list of candidate items (ingredients + distractors), constraining its output to a JSON schema. The model identifies and labels mention spans, providing attributes such as match type, confidence, and optional coreference links. Coreference resolution for pronouns and demonstratives (\emph{it, this, that, them}) is guided by nearby cooking-related actions.

\paragraph{Post-processing.} To ensure accuracy and consistency, we apply a series of deterministic rules: (1) normalization of predicted strings to canonical recipe or distractor names; (2) alias enrichment to handle synonyms (e.g., \emph{tap} $\rightarrow$ \emph{sink}); (3) precise alignment of timings and surface text; and (4) construction of a mention graph with unique IDs and antecedent links for coreferences. Automated checks are used to flag unmapped items, missing antecedents, or timestamp errors.

Full details on the prompt schema, processing rules, and handling of edge cases are documented in Appx.~\ref{app:prompt-schema}--\ref{app:linking-details}.

\subsection{Object detection and tracking.} 
\begin{figure}[t]
    \centering
        \includegraphics[width=0.8\linewidth]{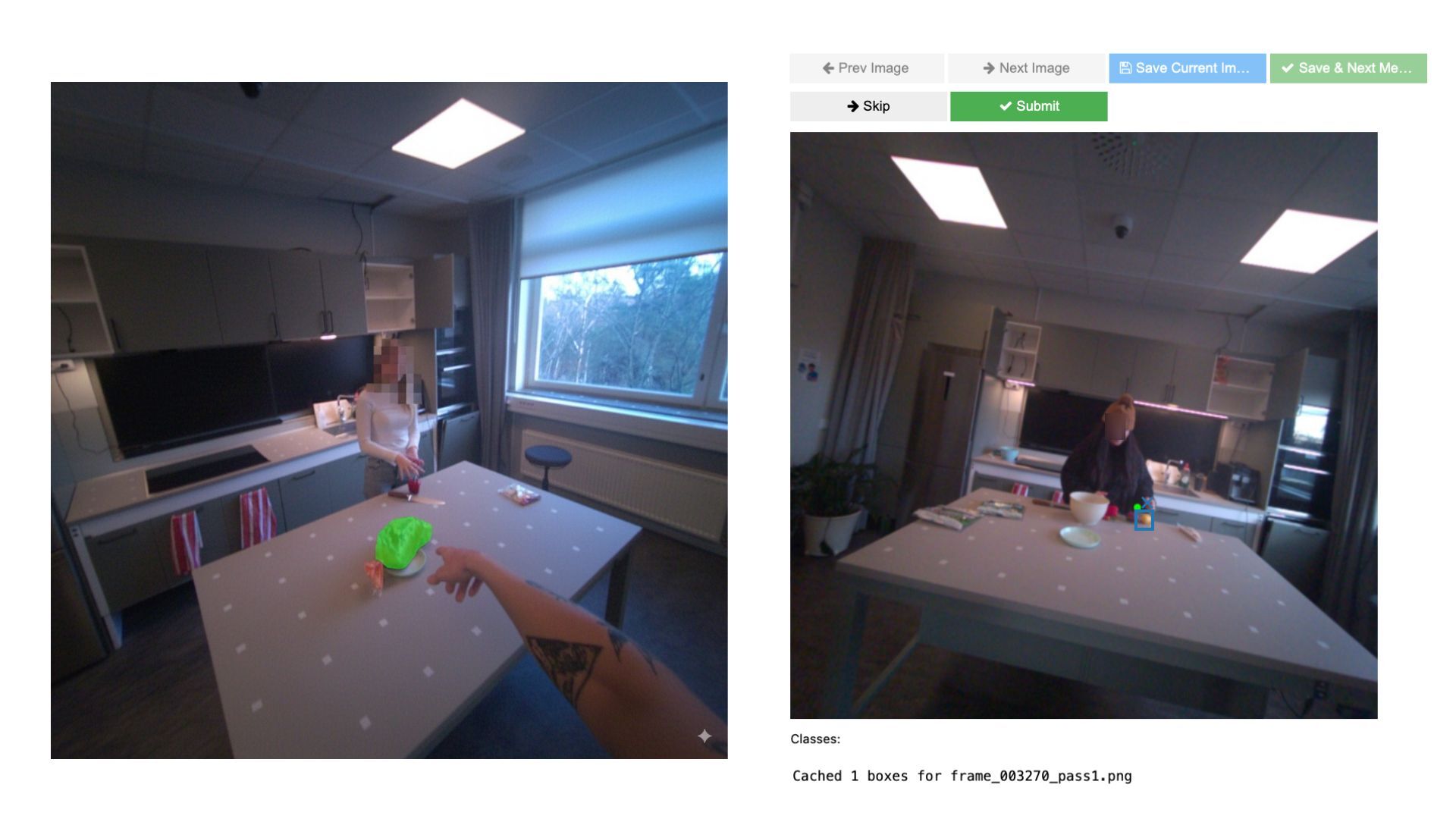}

    \caption{Examples of annotation strategies: (a) Molmo detection based automated SAM2 mask propagation, 
    (b)manual seeding of tracker with annotation interface. }
    \label{fig:annotation_examples}
\end{figure}

For each candidate mention, we applied \textbf{Molmo} \cite{deitke2024molmo}, a multimodal vision--language model, to localize the referenced item (ingredient or object; see definition above) in the undistorted frames.
We prompted Molmo with concise imperatives (e.g., \emph{``Point to the \{\texttt{description}\} object.''}) and evaluated it on frames sampled from the mention interval, using at least $\max(n_\text{frames}, n_\text{min})$ frames.
Molmo returned 2D point coordinates (in percent of image width/height), which we converted into seed locations to initialize the \textbf{SAM2}\cite{ravi2024sam} tracker.
SAM2 then propagated segmentation masks across the full mention interval, yielding dense per-frame masks aligned to each mention.
When Molmo did not yield a reliable localization, we performed \emph{manual annotation} to seed or correct the tracker, 
particularly for tiny items (e.g., spice containers) and visually similar look-alikes (e.g., sugar vs.\ wheat-flour jars; 
see Fig.~\ref{fig:annotation_examples}). 
In total, 747 mentions required manual handling: 106 cases were skipped because the object was not visible in the frames, 
while 641 mentions were manually annotated after inspection identified them as either too small (Molmo never reliable) 
or belonging to ambiguous object categories.

Together, the pipeline yields synchronized \textbf{speech transcripts}, \textbf{object mentions} (tokens $\rightarrow$ mentions $\rightarrow$ chains), \textbf{frame-level points}, \textbf{bounding boxes}, and \textbf{per-frame masks} for each referential episode.

\subsection{Analysis of Mention Annotations}

Across all 25 participants and 125 sessions, we annotated \textbf{2,707 mentions}
($\sim$22 per session). Table~\ref{tab:mention-types} shows the distribution:
ingredients dominate (62\%), while \textbf{23\% are pronouns or coreferential forms},
highlighting the prevalence of indirect reference. Objects (6\%) and distractors
($<2\%$) provide additional ecological variation.

Coreference chains averaged \textbf{6.7 mentions}, indicating repeated reference
to the same item once introduced. Ingredient coverage was high: nearly all
target items appeared at least once (mean coverage $>90\%$). 

\begin{table}[t]
\centering
\small
\setlength{\tabcolsep}{6pt}
\renewcommand{\arraystretch}{1.05}
\begin{tabular}{lrr}
\toprule
\textbf{Mention type} & \textbf{Count} & \textbf{Proportion} \\
\midrule
Ingredients      & 1,680 & 62.1\% \\
Pronoun/coref    &   614 & 22.7\% \\
Additional objects          &   154 &  5.7\% \\
Distractors      &    28 &  1.0\% \\
\midrule
\textbf{Total}   & 2,707 & 100\% \\
\bottomrule
\end{tabular}
\caption{Distribution of annotated mention types across the dataset.}
\label{tab:mention-types}
\end{table}

\begin{figure}[ht]
    \centering
    \begin{minipage}{0.32\textwidth}
        \centering
        \includegraphics[width=0.9\linewidth]{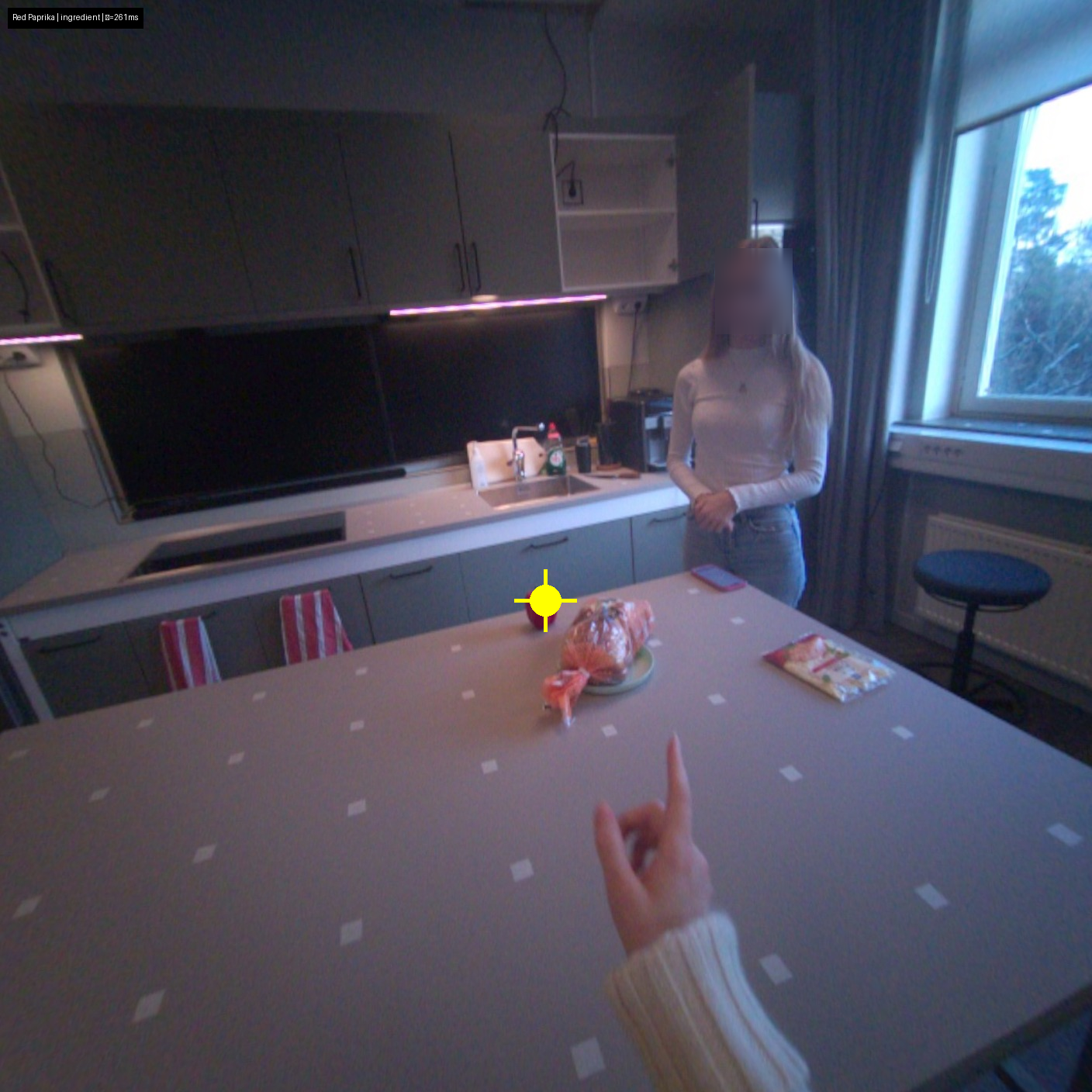}
        \caption*{Paprika mention \\ (“slice the paprika”)}
    \end{minipage}\hfill
    \begin{minipage}{0.32\textwidth}
        \centering
        \includegraphics[width=0.9\linewidth]{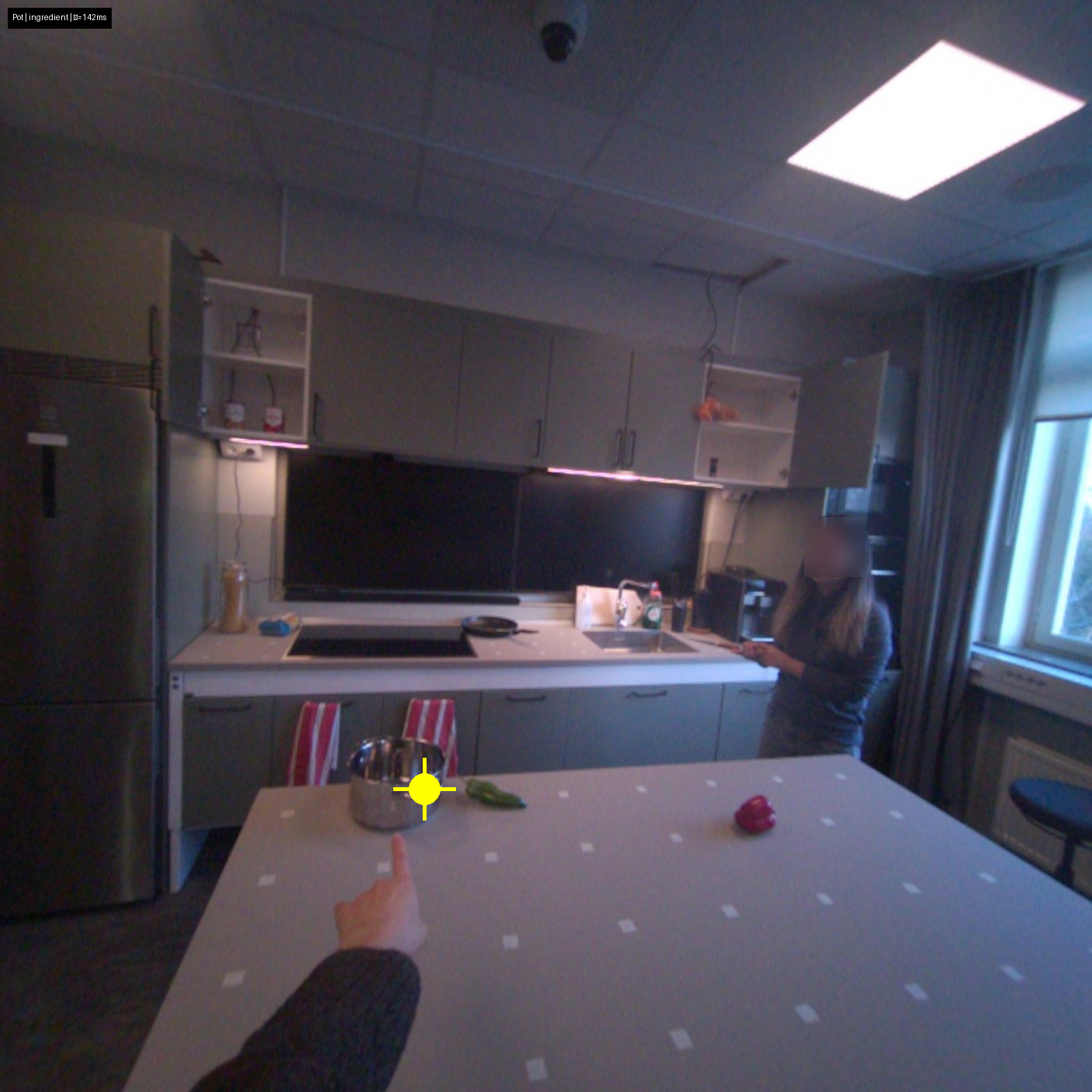}
        \caption*{Pot mention \\ (“put some water in it”)}
    \end{minipage}\hfill
    \begin{minipage}{0.32\textwidth}
        \centering
        \includegraphics[width=0.9\linewidth]{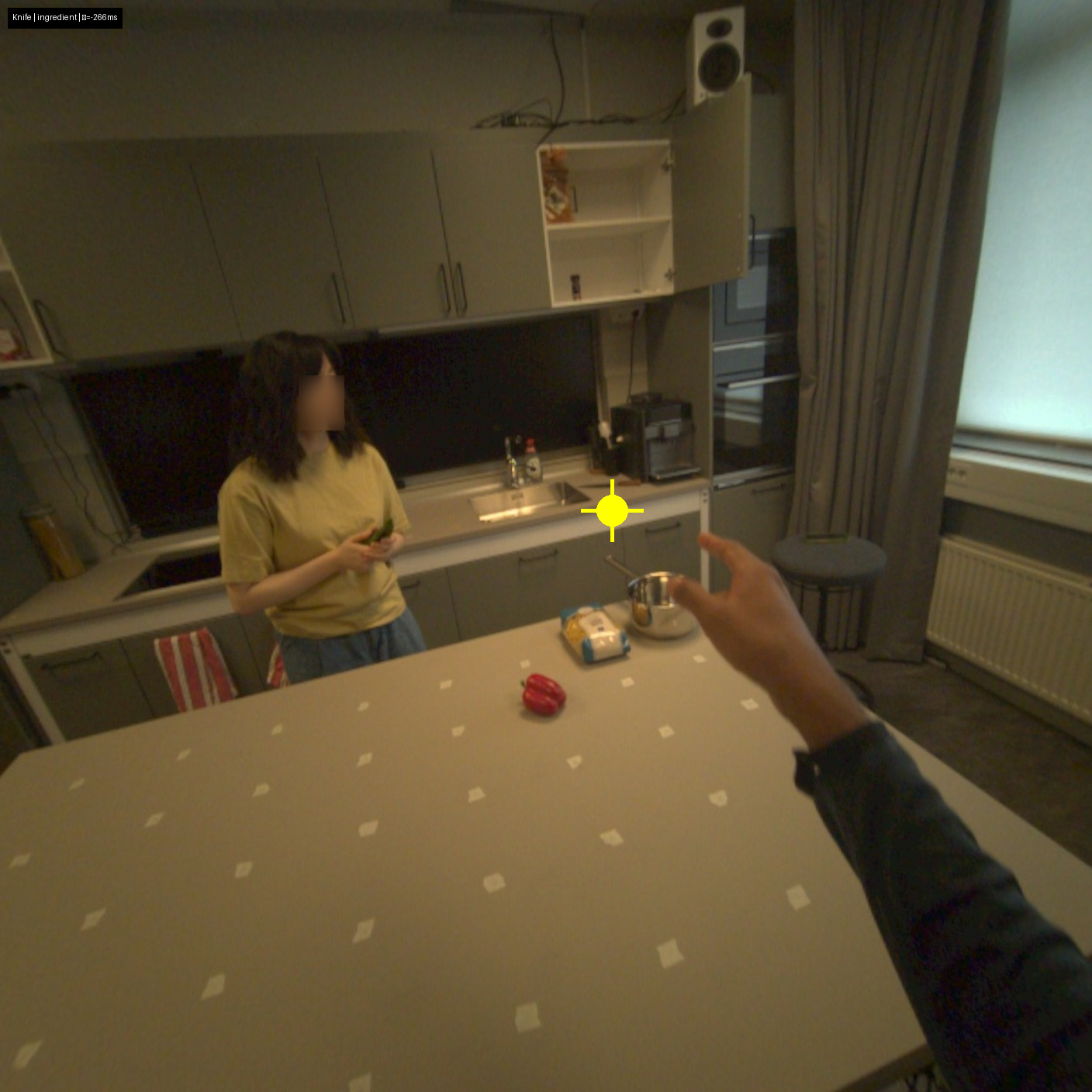}
        \caption*{Knife mention \\ (“grab the knife”)}
    \end{minipage}
    
    \caption{Examples of annotated mentions: (left) paprika, (middle) pot, (right) knife.}
    \label{fig:mentions}
\end{figure}

Although modest in scale, these annotations capture diverse referential behavior,
including explicit naming, pronouns, long chains, and variable timing—features
crucial for studying multimodal reference resolution.
Examples of annotated mentions are shown in Fig.~\ref{fig:mentions}.


\subsection{Gaze--Speech Synchrony Analysis}

\begin{figure}[h!]
    \centering
    \begin{minipage}[c]{0.45\linewidth}
        \vspace{0pt}\centering
        \includegraphics[width=\linewidth]{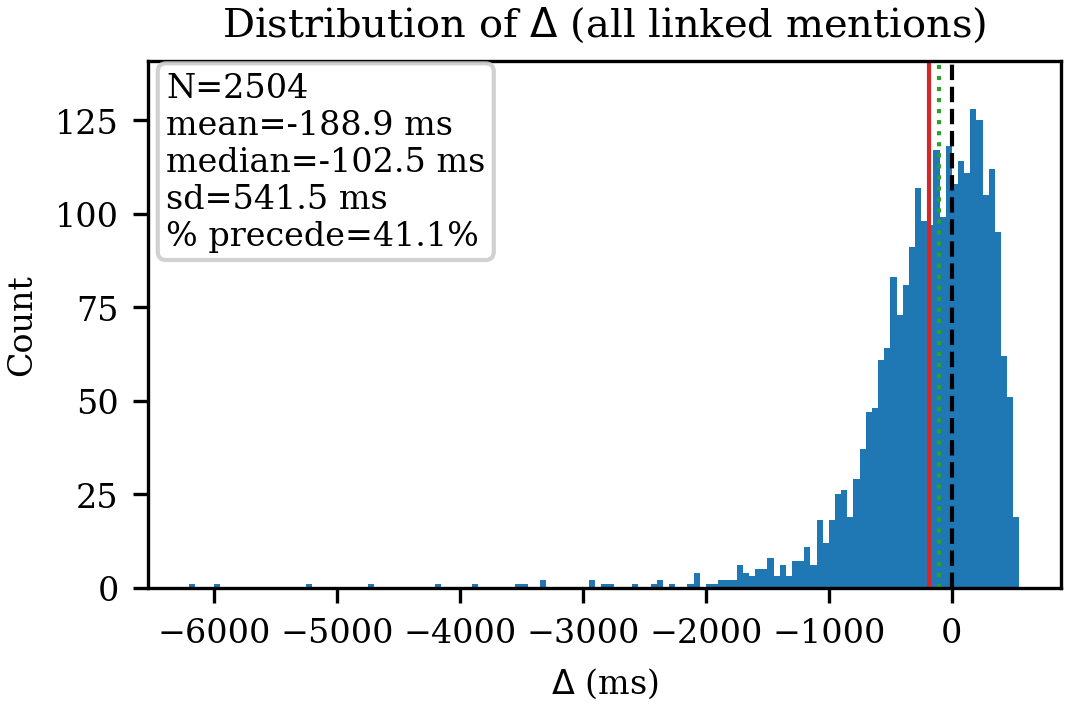}
        \caption*{(a) Distribution of gaze--speech lag $\Delta$ (fixation offset relative to mention onset). Positive values indicate gaze precedes speech.}
    \end{minipage}\hfill
    \begin{minipage}[c]{0.48\linewidth}
        \vspace{0pt}\centering
        \scriptsize
        \begin{tabular}{@{}lrr@{}}
        \toprule
        \textbf{Metric} & \textbf{Value} & \textbf{95\% CI} \\
        \midrule
        Total mentions      & 2707            & -- \\
        Linked mentions     & 2504 (92.5\%)   & -- \\
        $\Delta$ mean (ms)  & $-188.9$        & [$-210.7$, $-167.9$] \\
        $\Delta$ median (ms)& $-102.5$        & -- \\
        $\Delta$ SD (ms)    & 541.5           & -- \\
        $\Delta$ range (ms) & $[-6159, +531]$ & -- \\
        \% gaze precedes    & 41.1\%          & [39.2\%, 43.1\%] \\
        Overlap mean (ms)   & 351.9           & -- \\
        Overlap median (ms) & 366.6           & -- \\
        \bottomrule
        \end{tabular}
         \vspace{13pt}
        \caption*{(b) Summary statistics for gaze--speech synchrony.}
    \end{minipage}
    \caption{Gaze--speech synchrony: (a) distribution of lag $\Delta$ between mention onsets and fixation offsets, and (b) summary statistics.}
    \label{fig:gaze_speech}
\end{figure}

To quantify the temporal relationship between gaze and speech, we linked
2{,}504 mentions (92.5\% of all 2,707 mentions) to at least one fixation event. We define $\Delta$ as fixation offset minus mention onset. Figure~\ref{fig:gaze_speech}\,(a) shows the distribution of the time lag
$\Delta$ between mention onsets and fixation offsets, where
$\Delta > 0$ indicates gaze precedes speech.
The mean lag was $-189$\,ms (median $-102$\,ms, SD $=541$\,ms), with values
ranging from $-6159$\,ms to $+531$\,ms. Confidence intervals around the mean
were [$-211$, $-168$]\,ms.
In total, 41.1\% of mentions were preceded by gaze (95\% CI [39.2\%, 43.1\%]).
The average temporal overlap between gaze and mention was
$352$\,ms (median $367$\,ms).

These results indicate that while gaze and speech often overlap, gaze preceded the verbal mention in 41.1\% of instances. In this naturalistic task, this suggests that speakers frequently initiate an utterance while still fixating on the referent, rather than consistently shifting their gaze before speaking.


\section{Dataset Card and Ethical Considerations}

\textbf{Dataset Maintenance.} The Look and Tell dataset, including all annotations and supplementary materials, will be made publicly available upon publication. It will be hosted on Hugging Face Datasets \citep{lhoest-etal-2021-datasets} to ensure long-term accessibility. We commit to maintaining the dataset, addressing issues raised by the community, and providing a clear versioning system for any future updates, such as the gesture annotations mentioned in our future work. A website with documentation and tutorials will also be provided at: \url{https://huggingface.co/datasets/annadeichler/KTH-ARIA-referential}.

\textbf{Ethical Considerations.} Participants provided informed written consent for the recording of their video, audio, and gaze, and for the use of this data in anonymized form for research purposes. 

\textbf{Participant Privacy.} To protect participant privacy, all faces in the egocentric and exocentric video streams have been blurred using a standard face detection and blurring algorithm. While voices are not altered to preserve the speech data, we have reviewed the transcripts to ensure no personally identifiable information beyond what is inherent in the task was mentioned.

\textbf{Limitations and Bias.} Our dataset, while valuable, has limitations. The data was collected in a single kitchen environment (KTH Kitchen Lab), which may introduce environmental biases. The participant pool of 25 individuals, while diverse in native language, is primarily composed of university students and staff (aged 22-37), and may not be representative of the broader population in terms of age or background. Furthermore, the tasks are centered around specific recipes, which may favor individuals with some cooking familiarity. These factors should be considered when generalizing findings from this dataset.

\section{Discussion}
A core challenge in developing embodied AI is achieving robust multimodal spatial grounding, the ability to connect language to objects and locations within a shared 3D environment. Our dataset provides a unique opportunity to study how gaze, gestures, and speech unfold together in situated referential communication to achieve this. By uniquely combining egocentric (first-person) and exocentric (third-person) recordings with metadata supporting both 2D and 3D scene representations, our dataset offers a controlled \textbf{benchmark for evaluating spatial intelligence}. It is explicitly designed to facilitate research into \textbf{multimodal spatial grounding}, supporting analyses of how gaze can act as a predictive cue for resolving referential expressions and how different spatial representations may influence grounding performance. We hope this resource will foster the cross-disciplinary dialogue necessary to build embodied agents that can truly understand and communicate about space.

\textbf{Future Work.} The recorded sessions contain rich gestural behavior, including pointing and co-speech gestures, which are essential for multimodal reference resolution. Future extensions of this dataset will therefore include systematic gesture annotation, enabling analysis of how gaze, gesture, and speech jointly contribute to referential communication. This will allow embodied agents to be benchmarked not only on gaze-speech synchrony but also on their ability to integrate gestural cues into grounding and dialogue.

\section*{Acknowledgments}
This work was funded by the Digital Futures project \emph{Adaptive Intelligent Homes}. We thank Kristín Hafsteinsdóttir and Yu Lu for their assistance with data collection and annotation.

\bibliographystyle{plainnat} 
\bibliography{refs} 
\appendix

\section{Mention Linking Details}
\label{app:linking-details}

\subsection{Prompt Schema}
\label{app:prompt-schema}

The GPT model is constrained to output a JSON object with the following fields:

\begin{itemize}
  \item \texttt{start}, \texttt{len}: token index and length of the mention span
  \item \texttt{ingredient}: string or \texttt{null}
  \item \texttt{match\_type} $\in$ \{\texttt{exact}, \texttt{synonym}, 
        \texttt{hypernym}, \texttt{brand}, \texttt{coref}, \texttt{desc}, 
        \texttt{object}, \texttt{none}\}
  \item \texttt{confidence} $\in [0,1]$
  \item \texttt{antecedent\_start}, \texttt{antecedent\_len}, 
        \texttt{antecedent\_text} (optional, for coreference)
\end{itemize}

\subsection{Coreference Rules}
\label{app:coref-rules}

Pronouns and demonstratives (\emph{it, this, that, them, these, those})
are linked to the nearest plausible prior mention within approximately 25 tokens,
guided by cooking actions. Examples include:
\begin{itemize}
  \item \emph{open} $\rightarrow$ jar or can
  \item \emph{drain} $\rightarrow$ pasta or pot
  \item \emph{put on the stove} $\rightarrow$ pan or pot
  \item \emph{cut} $\rightarrow$ food item or package
\end{itemize}
Ambiguous cases default to \texttt{ingredient = null, match\_type = none}.

\subsection{Post-processing Steps}
\label{app:postproc}

\begin{enumerate}
  \item \textbf{Normalization:} canonicalize predicted strings to 
        recipe or distractor names; non-matches become \texttt{None}.
  \item \textbf{Enrichment:} add common object aliases 
        (e.g., \emph{hob/cooktop} $\rightarrow$ \emph{stove}) 
        and mark \texttt{is\_object}.
  \item \textbf{Timing and surface:} attach token indices and timestamps 
        (\texttt{start\_ns}, \texttt{end\_ns}, \texttt{start\_s}, \texttt{end\_s}) 
        and surface text.
  \item \textbf{Mention graph:} assign \texttt{mention\_id} and 
        \texttt{chain\_id}, infer \texttt{mention\_type} (including a small 
        utensil lexicon), and link \texttt{antecedent\_id} for coreference.
\end{enumerate}

\subsection{Automated Checks}
\label{app:checks}

We flag the following cases:
\begin{itemize}
  \item unmapped ingredients or objects,
  \item missing antecedents for coreference mentions,
  \item non-monotone or overlapping timestamps.
\end{itemize}

These rules ensure consistent and temporally grounded mentions for
each recording \texttt{ParXrecY}.

\end{document}